\documentclass[11pt]{article}

\usepackage{acl2012}
\usepackage{times}
\usepackage{linguex}
\usepackage{latexsym}
\usepackage{amsmath,bm}
\usepackage{multirow}
\usepackage{url}
\usepackage[hidelinks]{hyperref}
\usepackage{booktabs}
\usepackage{tikz}
\usepackage[round]{natbib}
\usepackage{dblfloatfix}
\usepackage{fixltx2e}
\usepackage{enumitem}
\setlist{noitemsep}

\usepackage{algorithm}
\usepackage[noend]{algpseudocode}

\usepackage{todonotes}

\usepackage{amsmath,amssymb, mathtools}
\usepackage{scalefnt}

\newcommand{\Cb}{\mathbf{C}}

\newcommand{\Eb}{\mathbf{E}}

\newcommand{\Mb}{\mathbf{M}}

\newcommand{\Zb}{\mathbf{Z}}

\newcommand{\cb}{\mathbf{c}}

\newcommand{\rb}{\mathbf{r}}
\newcommand{\ssb}{\mathbf{s}}

\newcommand{\pipe}{\;|\;}

\makeatletter
\newcommand*{\centerfloat}{%
  \parindent \z@
  \leftskip \z@ \@plus 1fil \@minus \textwidth
  \rightskip\leftskip
  \parfillskip \z@skip}
\makeatother

\title{Computational linking theory}

\author{\href{http://aswhite.net}{Aaron Steven White} \;\;\; \href{http://pages.jh.edu/~dreisin2/}{Drew Reisinger} \;\;\; \href{http://www.clsp.jhu.edu/people/graduate-students/rachel-rudinger/}{Rachel Rudinger} \\ \href{http://sites.krieger.jhu.edu/rawlins/}{\textbf{Kyle Rawlins}} \;\;\; \href{http://www.cs.jhu.edu/~vandurme/}{\textbf{Benjamin Van Durme}} \\ Johns Hopkins University}

\date{\today}

\allowdisplaybreaks

\begin{document}
\maketitle
\begin{abstract}
A \textit{linking theory} explains how verbs' \textit{semantic arguments} are mapped to their \textit{syntactic arguments}---the inverse of the \textit{semantic role labeling} task from the \textit{shallow semantic parsing} literature. In this paper, we develop the \textit{computational linking theory} framework as a method for implementing and testing linking theories proposed in the theoretical literature. We deploy this framework to assess two cross-cutting types of linking theory: \textit{local} v. \textit{global models} and \textit{categorical} v. \textit{featural models}. To further investigate the behavior of these models, we develop a measurement model in the spirit of previous work in \textit{semantic role induction}: the \textit{semantic proto-role linking model}. We use this model, which implements a generalization of \citeauthor{dowty_thematic_1991}'s seminal proto-role theory, to induce \textit{semantic proto-roles}, which we compare to those \citeauthor{dowty_thematic_1991} proposes.
\end{abstract}

\setlength{\Exlabelsep}{.1em}

\section{Introduction}
\label{sec:introduction}

\usetikzlibrary{bayesnet}

A \textit{linking theory} explains how verbs' \textit{semantic arguments} are mapped to their \textit{syntactic arguments} \citep{fillmore_grammar_1970,  zwicky_manner_1971, jackendoff_semantic_1972,  carter_linking_1976, pinker_language_1984, pinker_learnability_1989, grimshaw_argument_1990, levin_english_1993}. For example, the verb \textit{hit} has three semantic arguments---one for the \textsc{hitter}, one for the \textsc{hittee}, and one for the hitting \textsc{instrument}---and for each token of \textit{hit}, a subset of those semantic arguments are mapped to its syntactic arguments---e.g. subject, direct object, or object of a preposition.

\ex. \label{ex:good}
\a. [John]$_{\textsc{hitter}}$ hit [the fence]$_{\textsc{hittee}}$.
\b. [The stick]$_{\textsc{inst}}$ hit [the fence]$_{\textsc{hittee}}$.

\ex. \label{ex:bad}
\a. \#{}[The fence]$_{\textsc{hittee}}$ hit [John]$_{\textsc{hitter}}$.
\b. \#{}[The fence]$_{\textsc{hittee}}$ hit [the stick]$_{\textsc{inst}}$.

The main desideratum for selecting a linking theory is how well it explains \textit{linking regularities}: which mappings do and do not occur. One example of a linking regularity is that \textsc{hittee} arguments cannot be mapped to subject, suggested by the fact that \ref{ex:good} and \ref{ex:bad} cannot mean the same thing.

The task of constructing a linking theory that covers the entire lexicon is no small feat. One classic (though not the only) example of this difficulty concerns \textit{psych verbs}, like \textit{fear} and \textit{frighten} \citep{lakoff_irregularity_1970, postal_raising:_1974,perlmutter_1-advancement_1984, baker_incorporation:_1988, dowty_thematic_1991,pesetsky_zero_1995}. 

\ex.
\a. [Mary]$_{\textsc{fearer}}$ feared [John]$_{\textsc{fearee}}$.
\b. \#[John]$_{\textsc{fearee}}$ feared [Mary]$_{\textsc{fearer}}$. 

\ex.
\a. \#[Mary]$_{\textsc{fearer}}$ frightened [John]$_{\textsc{fearee}}$. 
\b. [John]$_{\textsc{fearee}}$ frightened [Mary]$_{\textsc{fearer}}$.

Psych verbs raise issues for theories that disallow mapping \textsc{fearer} to subject, since \textit{fear} does that, as well as those that disallow mapping \textsc{fearee} to subject, since \textit{frighten} does that.\footnote{See \citealt{hartshorne_love_2015} for recent work on how children learn psych verbs' linking regularities.}

Linking theory is intimately related to \textit{semantic role labeling} (SRL) \citep{gildea_automatic_2002,litkowski_senseval-3_2004,carreras_introduction_2004,marquez_semantic_2008}, which is a form of \textit{shallow semantic parsing}. Where a linking theory maps from semantic arguments to syntactic arguments, an SRL system maps from syntactic arguments to semantic arguments. Thus, SRL systems can be thought of as interpreting language, and linking theory implementations can be thought of as generating language.\footnote{See \citealt{flanigan_generation_2016} for recent work on semantics-based language generation with a looser coupling to the syntax.} But while much work has focused on building wide-coverage SRL systems, linking theory has not commanded similar attention.

In this paper, we introduce a framework for implementing linking theories---\textit{computational linking theory} (CLT)---which substantially generalizes an idea first introduced by \citet{grenager_unsupervised_2006}. In CLT, the traditional linking theoretic notion of a mapping from the space of semantic arguments $Sem$ to the space of syntactic arguments $Syn$ is implemented as a classifier.

This classifier can take several forms based on the structure of $Syn$. For instance, following \citet{lang_unsupervised_2010}, who build on \citealt{grenager_unsupervised_2006}, $Syn$ might be a set of syntactic positions, such as \{subject, object, $\dots$\}, in which case $Sem$ might be a set of thematic roles, such as \{\textsc{agent}, \textsc{patient}, $\dots$\}. Another possibility is that $Syn$ is a set of syntactic position sequences such as \{(subject, object), (subject, oblique), $\dots$\}, in which case $Sem$ might similarly be a set of thematic role sequences \{(\textsc{agent}, \textsc{patient}), (\textsc{agent}, \textsc{inst}), $\dots$\}, and the classifier would involve \textit{structured prediction}.

In the first part of this paper, we deploy CLT in conjunction with PropBank \citep{palmer_proposition_2005}, VerbNet \citep{kipper-schuler_verbnet:_2005}, SemLink \citep{loper_combining_2007}, and \citeauthor{reisinger_semantic_2015}'s (\citeyear{reisinger_semantic_2015}) recently released Semantic Proto-Roles version 1 (SPR1) dataset to evaluate the efficacy of various linking theories proposed in the theoretical literature. In the second part, we show that CLT can be useful not only for evaluation, but also for exploratory analysis, by developing a measurement model---the \textit{Semantic Proto-Role Linking Model} (\textsc{sprolim})---for analyzing computational linking theories. And though our main aim is to compare and explore theoretical proposals using computational tools, we believe that those interested in \textit{semantic role induction} (SRI) will find this measurement model useful for incorporating independent semantic annotations into SRI.

We focus on two cross-cutting types of linking theories that have been proposed in the theoretical literature: \textit{local} v. \textit{global models} and \textit{categorical} v. \textit{featural models}. The distinction between local and global models---which, as we discuss in \S\ref{sec:relatedwork}, is analogous to the distinction between local and global SRL systems \citep[cf.][]{toutanova_joint_2005, toutanova_global_2008}---contacts a long-standing theoretical debate regarding whether semantic arguments are mapped to syntactic positions independently of other arguments \citep{baker_incorporation:_1988} or whether there are dependencies among semantic arguments \citep{dowty_thematic_1991}. There is general consensus among theoreticians that this debate has been won in favor of localist theories---a consensus that we hope to break here.

The distinction between categorical and featural models contacts an independent debate as to whether semantic arguments, such as \textsc{hitter} and \textsc{hittee}, fall into discrete semantic role categories, such as \textsc{agent} or \textsc{patient}, or whether they are associated to a greater or lesser extent with fuzzy semantic role prototypes, such as \textsc{protoagent} and \textsc{protopatient} \citep{dowty_thematic_1991}. Because the featural models have been far less developed in the theoretical literature---largely due to lack of good methodologies for understanding their behavior---our goal here will be to further develop measurement models for exploring featural theories.

Our main findings are: 

\begin{enumerate}
\item Global models outperform local models (\S\ref{sec:categoricalfeatural})
\item Categorical models outperform featural models, particularly for oblique arguments (\S\ref{sec:categoricalfeatural})
\item \citeauthor{dowty_thematic_1991}'s \textsc{protoagent} prototype is robustly discovered by our measurement model, but his \textsc{protopatient} prototype appears to be a collection of multiple other prototype roles (\S\ref{sec:proto})
\end{enumerate} 

\noindent We begin with a discussion of related work in the statistical machine translation and shallow semantic parsing literatures, and we give a brief introduction to linking theory (\S\ref{sec:relatedwork}). We then describe the three datasets (PropBank, VerbNet, and SPR1) we build on to implement linking models (\S\ref{sec:data}). Based on these data, we implement and test four linking models built from crossing the categorical-featural distinction with the local-global distinction, and we establish the unequivocal superiority of the global models (\S\ref{sec:categoricalfeatural}). These experiments reveal challenges faced by the featural model, which we investigate using our \textit{Semantic Proto-Role Linking Model} (\S\ref{sec:proto}).  

\section{Related work}
\label{sec:relatedwork}

A \textit{linking theory} explains how verbs' \textit{semantic arguments} are mapped to their \textit{syntactic arguments}. Various types of theories have been proposed, differing mostly on how they define semantic roles. All of them share the feature that they predict syntactic position based on some aspect of the verb's semantics. 

\subsection{Predicting syntactic position}

The task of predicting an argument's syntactic position based on some set of linguistic features is not a new one in computational linguistics and natural language processing \citep[cf.][]{hajic_natural_2004}. This problem has been particularly important in the area of \textit{statistical machine translation} (SMT), where one needs to translate from morphologically poor languages like English to morphologically richer languages like Japanese and German \citep{koehn_europarl:_2005}. 

SMT researchers have focused for the most part on using morphological and syntactic predictors. \citet{suzuki_learning_2006,suzuki_generating_2007} construct models for predicting Japanese morphological case (which marks syntactic position in languages that have such cases) using intralanguage positional and alignment-based features, and \citet{jeong_discriminative_2010} extend this line of work to Bulgarian, Czech, and Korean. \citet{koehn_factored_2007}, \citet{avramidis_enriching_2008}, and \citet{toutanova_applying_2008} use richer phrase-based features to do the same task. 

Other approaches have incorporated semantic roles into SMT reranking components \citep{wu_semantic_2009}, similar to the reranking conducted in many SRL systems \citep[cf.][among others]{gildea_automatic_2002,pradhan_shallow_2004,pradhan_semantic_2005,pradhan_support_2005,toutanova_joint_2005,toutanova_global_2008}, but directly predicting syntactic position has not been explored in SMT (though see \citealt{minkov_generating_2007}, who suggest using semantic role information in future work).

\subsection{Semantic role labeling}

A semantic role labeling (SRL) system implements the inverse of a linking theory. Where a linking theory aims to map a verb's semantic arguments to it syntactic arguments, an SRL system aims to map a verb's syntactic arguments to its semantic arguments \citep{gildea_automatic_2002,litkowski_senseval-3_2004,carreras_introduction_2004,marquez_semantic_2008}. Continuing with examples \ref{ex:good} and \ref{ex:bad} from Section \ref{sec:introduction}, a linking theory would need to explain why (and when) \textsc{hitter}s and \textsc{instrument}s, but not \textsc{hittee}s, are mapped to subject position; in contrast, an SRL system would need to label the subject position with \textsc{hitter} or \textsc{instrument} (or some abstraction of those roles like \textsc{A}0 or \textsc{A}2) and the object with \textsc{hittee} (or some abstraction like \textsc{A}1).   

\paragraph{Local v. global models}

\citet{toutanova_joint_2005, toutanova_global_2008} introduce a distinction between local and global (joint) SRL models. In a local SRL model, a labeling decision is made based on only the features of the argument being labeled, while in a global system, features of the other arguments can be taken into account. The analogous distinction for a linking theory is between local linking models, which predict an argument's syntactic position based only on that argument's semantic role, and global linking models, which predict an argument's syntactic position based on its semantic role along with others'. In \S\ref{sec:categoricalfeatural}, we implement both local and global linking models for each representation of $Sem$ we consider.

\paragraph{Semantic role induction}

Semantic role annotation is expensive, time-consuming, and hard to scale. This has led to the development of unsupervised SRL systems for \textit{semantic role induction} (SRI). Work in SRI has tended to focus on using syntactic features to cluster arguments into semantic roles. \citet{swier_unsupervised_2004} introduce the first such system, which uses a bootstrapping procedure to first associate verb tokens with frames containing typed slots (drawn from VerbNet), then iteratively compute probabilities based on cooccurrence counts and fill unfilled slots based on these probabilities. 

\citet{grenager_unsupervised_2006} introduce the idea of predicting syntactic position based on a latent semantic role representation learned from syntactic and selectional features. \citet{lang_unsupervised_2010} expand on \citealt{grenager_unsupervised_2006} by introducing the notion of a \textit{canonicalized linking}. We discuss these ideas further in \S\ref{sec:proto}, incorporating both into our Semantic Proto-Role Linking Model (\textsc{sprolim}).

Syntax-based clustering approaches which do not explicitly attempt to predict syntactic position have also been popular. \citet{lang_unsupervised_2011, lang_similarity-driven_2014} use graph clustering methods and \citet{lang_unsupervised_2011-1} use a split-merge algorithm to cluster arguments based on syntactic context. \citet{titov_bayesian_2011} use a non-parametric clustering method based on the Pitman-Yor Process, and  \citet{titov_crosslingual_2012} propose two nonparametric clustering models based on the Chinese Restaurant Process (CRP) and distance dependent CRP. 

\subsection{Abstracting semantic roles}

To predict syntactic position, linking theories aim to take advantage of linking regularities. One way theories take advantage of linking regularities is to abstract over semantic arguments in such a way that the abstractions correlate with syntactic position. Two main types of abstraction have been proposed. On the one hand are categorical theories, which group semantic arguments into a finite set of \textit{semantic roles}---e.g., \textsc{hitter}s are \textsc{agent}s, \textsc{hittee}s are \textsc{patient}s, etc.---and then (deterministically) map these categories onto syntactic positions---e.g., subject, direct object, etc. \citep[][\textit{inter alia}; see \citealt{levin_argument_2005, williams_arguments_2015} for a review]{fillmore_grammar_1970,  zwicky_manner_1971, jackendoff_semantic_1972,  carter_linking_1976}. In a categorical theory, $Sem$ is thus some set of discrete indices, such as the core NP argument roles assumed in PropBank---i.e., $\{\text{{\sc A}0}, \text{{\sc A}1},\ldots\}$ \citep{palmer_proposition_2005}---or VerbNet---i.e., $\{\text{\sc agent}, \text{\sc patient},\ldots\}$ \citep{kipper-schuler_verbnet:_2005}.

On the other hand are featural theories, which assign each semantic argument a set of feature values based on predicate entailments imposed on that argument. For instance, \textsc{hitter}s are instigators and are thus assigned [+\textsc{instigates}]; they need not be volitional and are thus assigned [-\textsc{volitional}]; and they are not affected by the event and are thus assigned [-\textsc{affected}]; in contrast, \textsc{hittee}s are [-\textsc{instigates}], [-\textsc{volitional}], and [+\textsc{affected}]. A featural theory maps from (vectors of) those feature values onto syntactic positions. Thus, in a featural theory, $Sem$ is (or is related to) some set of vectors representing some priveleged set of $P$ entailments---e.g., $\{0, 1\}^P$, $\mathbb{R}^P$, etc.. A dataset that provides such a representation for verb-argument pairs in the Penn Treebank, which is also annotated for PropBank and (partially) for VerbNet roles, was recently made available by \citet{reisinger_semantic_2015}.\footnote{\url{http://decomp.net}}

\subsection{Role fragmentation}

Featural theories were initially proposed to remedy certain failings of the categorical theories. In particular, reasonably wide-coverage categorical theories require an ever-growing number of roles to capture linking regularities---a phenomenon that \citet{dowty_thematic_1991} refers to as \textit{role fragmentation}.    

In the face of role fragmentation, categorical theories have two choices: (i) use a large number of roles or (ii) force the fragmented roles into classes that predict the syntax well but map onto many distinct (possibly non-cohesive) semantic notions. These same choices must be made when developing a resource. For instance, FrameNet \citep{baker_berkeley_1998} makes the first choice; PropBank \citep{palmer_proposition_2005} makes the second; and VerbNet \citep{kipper-schuler_verbnet:_2005} splits the difference.

\begin{figure}
\includegraphics[scale=.23]{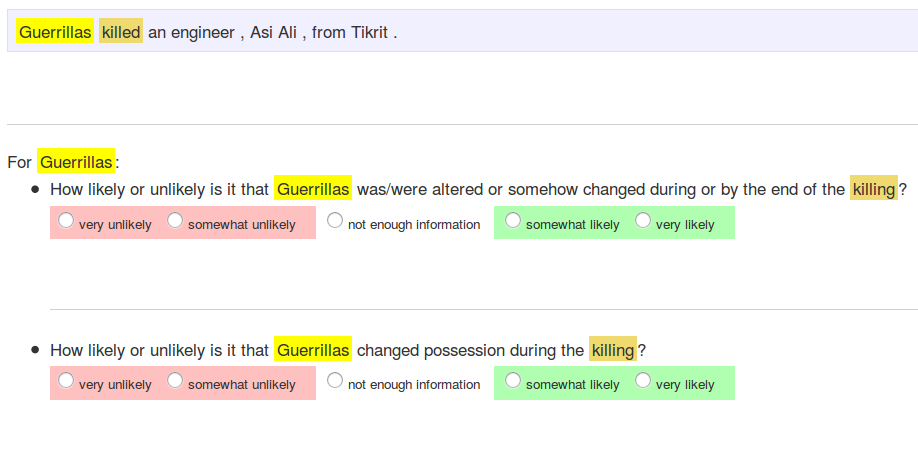}
\caption{Example of the SPR1 protocol}
\label{fig:spr1}
\vspace{-5mm}
\end{figure}

Featural theories remedy the role fragmentation found in categorical theories by positing a small number of properties that a verb might entail about each of its arguments. The properties that hold of a particular argument determine which position it gets mapped to. Different configurations of these properties correspond to a large space of roles. For example, $P$ binary properties generate $2^P$ potential roles. \citet{dowty_thematic_1991} proposes the first (and, perhaps, most comprehensive) list of such properties as a part of specifying his \textit{proto-role linking theory} (PRLT).  

In PRLT, properties are grouped into two clusters: \textsc{protoagent} properties and \textsc{protopatient} properties. These groupings are named this way for the fact that \textsc{agent}s in categorical theories, which tend to get mapped to subject position, tend to have \textsc{protoagent} properties and \textsc{patient}s, which tend to get mapped to a non-subject position, tend to have \textsc{protopatient} properties. With this idea in mind, \citeauthor{dowty_thematic_1991} proposed that semantic arguments were linked to syntactic arguments by (i) counting the number of \textsc{protoagent} and \textsc{protopatient} properties each argument has and (iia) mapping the argument with the most \textsc{protoagent} properties to subject or, (iib) if there is a tie for  \textsc{protoagent} properties, mapping the argument with the least \textsc{protopatient} properties to subject.

An important thing to note before moving on is that, unlike other linking theories, which are concerned with determining more general linking phenomena \citeauthor{dowty_thematic_1991} concerns himself only with determining which semantic argument is mapped to subject, though he does briefly suggest that his system might extend to selecting which argument is mapped to, e.g., object v. oblique position. We assess such an extension in the experiments described below.

\section{Data}
\label{sec:data}

For all experiments, datasets are based on the thematic role annotations of the Penn Treebank \citep{marcus_building_1993} found in PropBank \citep{palmer_proposition_2005} and VerbNet \citep{kipper-schuler_verbnet:_2005}, mapped via SemLink \citep{loper_combining_2007}, as well as the Semantic Proto-Roles version 1.0 (SPR1) dataset of crowd-sourced proto-role property annotations \citep{reisinger_semantic_2015}. 

The SPR1 annotations consist of answers to simple questions about how likely, on a five-point scale, it is that particular relational properties hold of arguments of PropBank-annotated verbs \citep[cp.][]{kako_thematic_2006, greene_more_2009,hartshorne_verbcorner_2013}. \citeauthor{reisinger_semantic_2015} constructed the questions to correspond to each of Dowty's proto-role properties. Figure \ref{fig:spr1} shows an example of the protocol. 

We extracted syntactic position information about each SPR1 annotated argument by mapping Penn Treebank annotations to Universal Dependencies \citep{de_marneffe_universal_2014,nivre_universal_2015} using the Stanford Dependencies to Universal Dependencies converter available in \texttt{PyStanfordDependencies}.\footnote{\url{https://pypi.python.org/pypi/PyStanfordDependencies}}

\subsection{Argument and clause filtering}

SPR1 contains only a subset of PTB sentences produced by applying various automated filters (see \citealt{reisinger_semantic_2015} for details). All of our models, including the PropBank- and VerbNet-based models are trained on this subset---or in the case of the VerbNet-based models, a subset thereof (see \S\ref{sec:verbnetpreprocess}). The most important of these filters for current purposes is one that retains only NP core arguments. This is useful here, since linking theories tend to only treat semantic arguments that surface as NPs.
In spite of these filters, some non-NPs occur in SPR1. To remedy this, we filter all dependents that do not have a UD dependency label matching \verb|nsubj|, \verb|dobj|, \verb|iobj|, or \verb|nmod|. 
\begin{table}
\centering
\begin{tabular}{lrr}
\toprule
Argument type &   SPR1.0  &  VN subset \\
\midrule
subject            &          5041 & 2357\\
object (direct)    &          2902 & 1307\\
oblique            &          1253 &  382 \\
object (indirect)  &            28 & 18 \\
\bottomrule
\end{tabular}
\caption{Count of each argument type after preprocessing \label{tab:gramfuncudcount}}
\begin{tabular}{lrr}
\toprule
Clause type    &   SPR1.0  &  VN subset \\
\midrule
NP V NP        &            2342 &            1144  \\
NP V           &            1508 &            823   \\
NP V PP+        &             625 &             221   \\
NP V NP PP+     &             544 &             157   \\
NP V NP NP     &              22 &               12   \\
\bottomrule
\end{tabular}
\caption{Count of each clause type after preprocessing \label{tab:sentencetypecount}}
\vspace{-5mm}
\end{table}

\subsection{VerbNet subset}
\label{sec:verbnetpreprocess}

VerbNet contains role annotations for only a subset of the SPR1 data. This has to do with the fact that only a subset of the verbs in the PTB that are annotated with PropBank rolesets are also annotated with VerbNet verb classes. A further subset of these verbs also have their arguments (whose spans are defined by PropBank) annotated with VerbNet semantic roles. Thus, there are three kinds of verbs in the corpus: those with no VerbNet annotation, those annotated only with verb class, and those with both verb class and semantic role annotations. We apply our models both to the full set of SPR1 annotations as well as the VerbNet-annotated subset.

\subsection{Final datasets}

Table \ref{tab:gramfuncudcount} gives the counts for each syntactic position after the preprocessing steps listed above, and Table \ref{tab:sentencetypecount} gives the analogous counts for each clause type. The argument type counts show only subjects (\verb|nsubj|), direct objects (\verb|dobj|), indirect objects (\verb|iobj|), and obliques (\verb|nmod|) because all other arguments are filtered out as non-core arguments. 

There are few indirect objects---i.e., first direct objects in a double object construction---in this dataset. This is problematic for the cross-validation we employ in \S\ref{sec:categoricalfeatural}. To remedy this, we collapse the indirect object label to the direct object label. This is justified linguistically, since indirect objects in the sense employed in Universal Dependencies (and most other dependency parse standards) is really just a subtype of direct object.

\section{Evaluating linking models}
\label{sec:categoricalfeatural}

In this section, we implement categorical and featural linking models by constructing classifiers that predict the syntactic position (\textit{subject}, \textit{direct object}, \textit{oblique}) of an argument based either on that argument's thematic role (e.g. \textsc{agent}, \textsc{patient}, etc.) or on the entailments that the verb requires of that argument (e.g. \textsc{instigation}, \textsc{volition}, etc.). 

For each type of predictor, two linear classifiers are constructed to instantiate (i) a \textit{local linking model} (Experiment 1), which predicts syntactic position irrespective of other arguments, and (ii) a \textit{global linking model} (Experiment 2), which predicts syntactic position relative to other arguments. In both experiments, the PropBank- and SPR1-based models are fit to the full SPR1 dataset, and all three models are fit to the VerbNet-annotated subset.

The featural models---in particular, the global featural model---can be seen as a generalization of \citeauthor{dowty_thematic_1991}'s proto-role model. Like \citeauthor{dowty_thematic_1991}'s model, it groups properties based on how predictive they are of particular syntactic positions---e.g., \textsc{protoagent} properties are predictive of subject position. They are a generalization in two senses: (i) for \citeauthor{dowty_thematic_1991}, each role is weighted equally---one simply counts how many of each kind of property hold---while here, these properties can receive distinct weights (and are ordinal- rather than binary-valued); and (ii) instead of predicting only subject v. non-subject; we attempt to also differentiate among non-subjects---i.e., direct object v. obliques. another way of thinking about the featural models is that they, like \citeauthor{dowty_thematic_1991}'s model, admit of role prototypes with piecewise linear boundaries in the property space.  

The main findings in this section are that (i) the global models substantially improve upon the local models for both categorical and featural predictors and (ii) the featural models perform worse overall than the categorical models. The first finding argues against local linking theories like that proposed by \citet{baker_incorporation:_1988}. We hypothesize that the second finding has two sources: (i) the set of properties in SPR1, which are essentially just \citeauthor{dowty_thematic_1991}'s properties, is insufficient for capturing distinctions among non-subject positions like \textit{direct object} and \textit{oblique}---likely because \citeauthor{dowty_thematic_1991} engineered his properties only to distinguish subjects from non-subjects; and (ii) because the models we use don't capture multimodality in the kinds of property configurations that exist.  We explore this second possibility in \S\ref{sec:proto}.

\subsection{Classifiers}

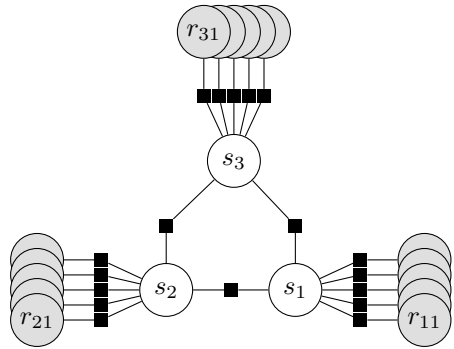
\begin{figure}
\centering
%
%
%
%


\begin{tikzpicture}[x=1cm,y=1cm]

  \node[latent]                  (arg1)      {$s_1$} ; %
  \node[latent, left=of arg1]       (arg2)      {$s_2$} ; %
  \node[latent, above=of arg2, xshift=.9cm]       (arg3)      {$s_3$} ; %

  
  
  \factor[left=of arg1] {a1a2-f} {} {arg1,arg2} {} ;
  \factor[above=of arg1] {a1a3-f} {} {arg1,arg3} {} ;
  \factor[above=of arg2] {a2a3-f} {} {arg2,arg3} {} ;
  
  \node[obs, right=of arg1, yshift=.4cm]       (feat1)      {} ; %
  \factor[left=of feat1] {f1a1-f} {} {feat1,arg1} {} ;
  
  \node[obs, right=of arg1, yshift=.2cm]       (feat2)      {} ; %
  \factor[left=of feat2] {f2a1-f} {} {feat2,arg1} {} ;
  
  \node[obs, right=of arg1]       (feat3)      {} ; %
  \factor[left=of feat3] {f3a1-f} {} {feat3,arg1} {} ;

  \node[obs, right=of arg1, yshift=-.2cm]       (feat4)      {} ; %
  \factor[left=of feat4] {f4a1-f} {} {feat4,arg1} {} ;

  \node[obs, right=of arg1, yshift=-.4cm]       (feat5)      {$r_{11}$} ; %
  \factor[left=of feat5] {f5a1-f} {} {feat5,arg1} {} ;

  \node[obs, left=of arg2, yshift=.4cm]       (feat1)      {} ; %
  \factor[right=of feat1] {f1a1-f} {} {feat1,arg2} {} ;
  
  \node[obs, left=of arg2, yshift=.2cm]       (feat2)      {} ; %
  \factor[right=of feat2] {f2a1-f} {} {feat2,arg2} {} ;
  
  \node[obs, left=of arg2]       (feat3)      {} ; %
  \factor[right=of feat3] {f3a1-f} {} {feat3,arg2} {} ;

  \node[obs, left=of arg2, yshift=-.2cm]       (feat4)      {} ; %
  \factor[right=of feat4] {f4a1-f} {} {feat4,arg2} {} ;

  \node[obs, left=of arg2, yshift=-.4cm]       (feat5)      {$r_{21}$} ; %
  \factor[right=of feat5] {f5a1-f} {} {feat5,arg2} {} ;

  \node[obs, above=of arg3, xshift=.4cm]       (feat1)      {} ; %
  \factor[below=of feat1] {f1a1-f} {} {feat1,arg3} {} ;
  
  \node[obs, above=of arg3, xshift=.2cm]       (feat2)      {} ; %
  \factor[below=of feat2] {f2a1-f} {} {feat2,arg3} {} ;
  
  \node[obs, above=of arg3]       (feat3)      {} ; %
  \factor[below=of feat3] {f3a1-f} {} {feat3,arg3} {} ;

  \node[obs, above=of arg3, xshift=-.2cm]       (feat4)      {} ; %
  \factor[below=of feat4] {f4a1-f} {} {feat4,arg3} {} ;

  \node[obs, above=of arg3, xshift=-.4cm]       (feat5)      {$r_{31}$} ; %
  \factor[below=of feat5] {f5a1-f} {} {feat5,arg3} {} ;

\end{tikzpicture}
\caption{Factor graph for global model applied to a three argument sentence. Variable $s_i$ represent the syntactic position of argument $i$. Variable $r_{ip}\equiv (a_{ip}, l_{ip})$ represents the likelihood $l_{ip}$ that property $p$ applies to argument $i$ and the applicability $a_{ip}$ of property $p$ to arg $i$. \label{fig:crf}}
\vspace{-7mm}
\end{figure}

\begin{table*}[t]
\centering
{\small
\begin{tabular}{lll|ccc|ccc|ccc}
\toprule
       &      &   & \multicolumn{3}{c}{PropBank} & \multicolumn{3}{c}{SPR1}  & \multicolumn{3}{c}{VerbNet}       \\
       &      &   &           F1 & precision & recall &       F1 & precision & recall &          F1 & precision & recall \\
\midrule
local & full & subject &         0.87 &      1.00 &   0.76 &     0.82 &      0.83 &   0.82 &          &        &     \\
       &      & object &         0.75 &      0.65 &   0.89 &     0.68 &      0.58 &   0.81 &          &        &     \\
       &      & oblique &         0.73 &      0.71 &   0.76 &     0.09 &      0.53 &   0.05 &          &        &     \\
       & subset & subject &         0.89 &      1.00 &   0.81 &     0.85 &      0.87 &   0.83 &        0.88 &      0.90 &   0.87 \\
       &      & object &         0.77 &      0.68 &   0.88 &     0.72 &      0.63 &   0.84 &        0.72 &      0.69 &   0.75 \\
       &      & oblique &         0.65 &      0.61 &   0.70 &     0.07 &      0.40 &   0.04 &        0.56 &      0.60 &   0.54 \\
global & full & subject &         0.92 &      0.92 &   0.91 &     0.91 &      0.91 &   0.90 &          &        &     \\
       &      & object &         0.83 &      0.78 &   0.89 &     0.77 &      0.71 &   0.85 &          &        &     \\
        &      & oblique &         0.73 &      0.88 &   0.63 &     0.48 &      0.66 &   0.38 &          &        &     \\
        & subset & subject &         0.89 &      0.89 &   0.89 &     0.85 &      0.85 &   0.85 &        0.87 &      0.87 &   0.87 \\
        &        & object &         0.81 &      0.76 &   0.88 &     0.77 &      0.73 &   0.82 &        0.78 &      0.73 &   0.83 \\
        &      & oblique &         0.54 &      0.84 &   0.39 &     0.12 &      0.17 &   0.09 &        0.50 &      0.75 &   0.38 \\
\bottomrule
\end{tabular}
}

\caption{Mean F1, precision, and recall on outer cross-validation test folds for local (Exp. 1) and global (Exp. 2) models on both full SPR1 dataset and the VerbNet subset. The VerbNet models were only run on the subset. \label{tab:exp1and2metrics}}
\end{table*}

L2-regularized maximum entropy models were used for classification in both Experiments 1 and 2. Experiment 1 uses simple logistic regression, and Experiment 2 uses a conditional random field (CRF) analogous to the logistic regression used for Experiment 1, but containing factors for each pair of arguments (see Figure \ref{fig:crf} for a three argument example).

\paragraph{Experiment 1}

Syntactic position $s_{i} \in $ \{\textit{subject}, \textit{object}, \textit{oblique}\} was used as the dependent variable and either thematic role or property configuration as predictors. Entailment judgments from SPR1 were represented as a vector $\mathbf{l}_i$ of likelihood ratings $l_{ij} \in \{1, 2, 3, 4, 5\}$ for each potential entailment $j$. 

Ratings $l_{ij}$ in SPR1 are furthermore associated with values $a_{ij} \in \{0, 1\}$ corresponding to the applicability of a particular entailment question. If a question $i$ was annotated as not applicable ($a_{ij}=0$), the combined rating $r_{ij}$ was set to $a_{ij}l_{ij} = 0$. Because $l_{ij}$ is strictly positive, by setting these ratings to 0, the classifier is effectively forced to project a probability from only the feature subspace corresponding to the applicable questions.

\paragraph{Experiment 2}

The sequence of syntactic positions in each clause was used as the dependent variable. For instance, \Next would be labeled \{\textit{subj}, \textit{obj}, \textit{obl}\}.

\ex. [The bill]$_{subj}$ also \textbf{imposes} [the California auto-emissions standards]$_{object}$ on [all cars nationwide]$_{oblique}$.

These sequences were predicted using the CRF corresponding to the factor graph in Figure \ref{fig:crf}. Because the maximum number of core NP arguments and syntactic position types for any verb token is relatively small, exact inference for $\mathbf{s}$ is possible in our case by enumerating all configurations in the relevant cartesian product of syntactic positions $\mathcal{S}$---$\mathcal{S}^2$ for a two-argument verb, $\mathcal{S}^3$ for a three-argument verb, etc.---and computing their probabilities explicitly.\footnote{We suspect that this strategy will generally be possible since the vast majority of verbs have only one or two core arguments (see Table \ref{tab:sentencetypecount}). Furthermore, even if the number of arguments were high on average, most configurations are not possible---e.g., one never finds two subjects. These sorts of syntax-aware constraints can significantly cut down the space of configurations that need to be considered.}

\subsection{Cross-validation}
\label{sec:exp1cv}

Nested stratified cross-validation with 10-folds at both levels of nesting was used to validate and test the models in both experiments. For each of the 10 folds in the outer CV, the L2 regularization parameter $\alpha$ was set using grid search over $\alpha \in \{0.01, 0.1, 1, 2, 5, 10\}$ on the 10 folds of the inner CV with highest average F1 as the selection criterion. For each of the outer folds, the model with this optimal $\alpha$ was refit to the full training set on that fold and tested on that fold's held-out data. In Experiment 2, a further constraint was imposed that folds not separate arguments in the same sentence. 

All reported F1, precision, and recall values are computed from testing on the outer held-out sets, and all error analyses are conducted on errors when an item was found in an outer held-out set. 

\subsection{Results}

Table \ref{tab:exp1and2metrics} gives the mean F1, precision, and recall on the outer cross-validation test folds on both full SPR1 dataset and the VerbNet subset.  There are two relevant patterns. First, across all predictor sets, the three measures for the global models improve substantially compared to the local models. This suggests that, regardless of one's representation of argument semantics, global models are to be preferred over local models. It furthermore suggests that the current trend in theoretical linguistics to prefer local models likely needs to be reassessed.

Second, we find that the featural models do worse than the categorical models, particularly comparing the respective local models, but that this gap is closed to some extent when considering the respective global models. This change is particularly apparent for obliques, for which the local featural model's performance is abysmal and the global featural model's performance is middling, at least when validated on the full dataset.\footnote{There is a substantial decrement in F1 for the obliques when comparing the global featural model on the full dataset and VerbNet subset. This seems to arise from higher confusion with subject. We address this in the error analysis.}   As such, in the remainder of this section, we focus in on understanding the behavior of the global models. 

\begin{figure}[t]
\input{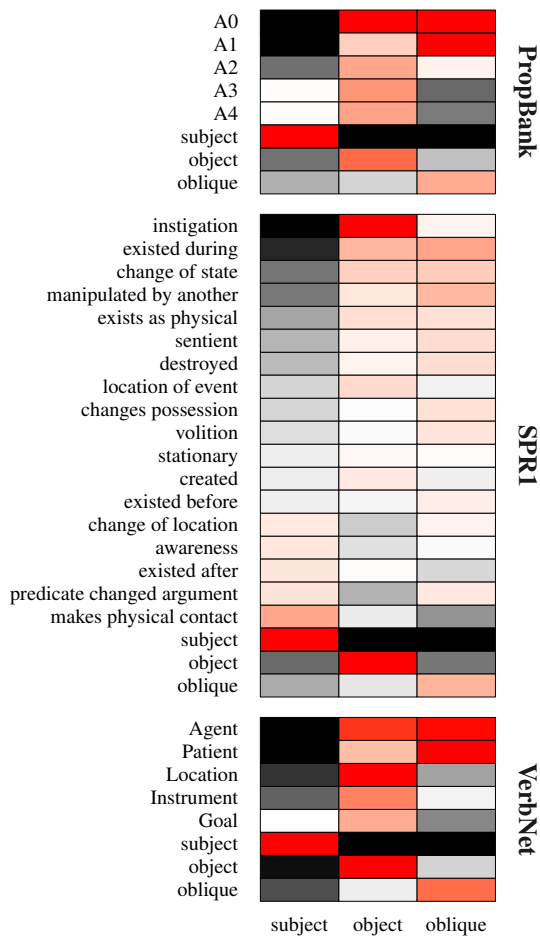}
\vspace{-2mm}
\caption{Heatmap of mean coefficients for global models. Black is +; red is --. \label{fig:coefs}}
\vspace{-5mm}
\end{figure}

Because these models are fairly simple, it is straightforward to analyze how they make their predictions by looking at their parameters. Figure \ref{fig:coefs} shows a heatmap of the mean coefficients for each global model across the 10 outer CV folds.

Turning first to the coefficients relating the syntactic positions to each other, we see that, across all models, each syntactic position disprefers occurring with a syntactic position of the same kind, and this dispreference is particularly strong for subjects. This makes sense in that we never find a sentence with two subjects, and duplicates of the other syntactic positions are relatively rare, only occurring in double object constructions or sentences with multiple prepositions. On the other hand, unlike syntactic positions attract, with strongest preference for sentences containing a non-subject to have a subject.\footnote{This is likely due to the fact that English requires subjects; in a language that has no such requirement, we might expect a different pattern.}

Turning next to the categorical roles, we see that the majority of roles either prefer subject position or are agnostic, but none disprefer it. In contrast, all disprefer object to some extent. This likely arises because each of these roles can occur in intransitives, which always place their single argument in subject position.\footnote{There are languages that show, e.g., distinct case-marking behavior for different intransitives \citep[][see also \citealt{stevenson_automatic_1999,stevenson_automatic_1999-1}]{perlmutter_impersonal_1978,burzio_italian_1986,levin_unaccusativity:_1995,hale_prolegomena_2002}, and so we might again expect a different pattern in those languages.} This pattern also gives rise to an ordering on the roles with respect to which role will be mapped to subject position when others are present---e.g., \textsc{A0} and \textsc{agent} will be mapped to subject before any other role. This is reminiscent of popular proposals from the theoretical literature regarding \textit{role hierarchies} (see \citealt{levin_argument_2005} for a review), and errors arise when that ordering is violated. For instance, because A2 does not disprefer objects as much as A1, (A1, A2) is mapped to (\textit{object}, \textit{subject}) by the model, when they should be mapped to (\textit{subject}, \textit{oblique}). 

\ex. [Approximately 85\% of the total]$_\text{A1}$ consisted [of nonperforming commercial real estate assets]$_\text{A2}$.

The featural model coefficients are slightly harder to interpret. We see that being likely to \textsc{instigate}, one of \citeauthor{dowty_thematic_1991}'s \textsc{protoagent} properties, matters a lot for being mapped to subject position and not being mapped to object, but the rest of the relationships are quite weak and their relationship with subject position doesn't match well with the predictions of a featural theory such as \citeauthor{dowty_thematic_1991}'s. For instance, \textsc{change of state} is supposed to be a \textsc{protopatient} property under his theory, but here it has its highest weight on subject. This appears to produce problems for many of the same sentences that the categorical models fail on---the featural model fails on 75\% of the sentences the PropBank model fails on and 64\% of the sentences the VerbNet model fails on---but it also produces problems for psych verbs and their kin---e.g., \Next---whose subjects are traditionally referred to as \textsc{experiencers}.   

\ex. The real-estate market suffered even more severe setbacks.

The likely reason for this is that, as for the categorical models, the featural model must capture the fact that subjects of intransitives can be arguments that might occur in subject position in the presence of a `better' subject. This might be further worsened by the existence of distinct clusters of properties that the model has no way of capturing. But because this representation doesn't provide a sense for which of the $6^{18}$ different combinations are extant, it is hard to tell what kinds of role combinations there are and, thus, which fall into each category. We could obtain a rough estimate of this by taking some statistic over the arguments that are classified as \textit{subject}, \textit{object}, and \textit{oblique}, but this would fail to capture categories with multimodality in the property space. If fully spelled out categorical theories are even a good approximation, one would expect such multimodality.

\subsection{Discussion}

In this section, we established (i) that global models substantially improve upon the local models for both categorical and featural predictors and (ii) that featural models perform worse overall than the categorical models. In the next section, we demonstrate CLT's use as a framework for exploring linking theories---particularly, featural linking theories---by developing a measurement model that addresses the multimodality issue raised in this section.

\section{Exploring linking models}
\label{sec:proto}

In this section, we present the Semantic Proto-Role Linking Model (\textsc{sprolim}), which is a multi-view mixture model for inducing \textit{semantic proto-roles} for each argument of a predicate from the property judgments employed in the last section. This model can be seen as a further generalization of \citeauthor{dowty_thematic_1991}'s proto-role theory that incorporates the idea that semantic roles have a prototype structure.  

In Experiment 3, we apply \textsc{sprolim} to the SPR1 data with the aim of discovering semantic protoroles. We investigate how the structure of the semantic protoroles changes as we allow for more distinct types of protoroles, finding that the one constant prototype is a \textsc{protoagent} role with exactly the structure proposed by \citeauthor{dowty_thematic_1991}. In contrast, \citeauthor{dowty_thematic_1991}'s \textsc{protopatient} role appears to rather be a collection multiple other protoroles.

\subsection{Semantic Proto-Role Linking Model}

There are four main components of \textsc{sprolim}. The first component is a representation of the relationship between a predicate's \textit{l(exical)-thematic roles} \citep{dowty_semantic_1989}---e.g., for the verb \textit{hit}, the \textsc{hitter}, \textsc{hittee}, and \textsc{hitting instrument} roles---and generalized \textit{semantic proto-roles}---e.g., \textsc{protoagent} and \textsc{protopatient}. To make clear that \textit{l-thematic roles} are abstract, we refer to them by the more transparent name \textit{argument types} and denote the set of argument types for a verb $v$ with $\mathcal{A}_v$.

The second component of \textsc{sprolim} is a representation of the relationship between the semantic proto-role that an argument has and (i) the likelihood that a property is applicable to that argument and, (ii) if applicable, how likely it is that the property holds of the argument. We call this second component the \textit{property model}. The property model represents this relationship probabilistically---i.e., each semantic proto-role is associated with a distribution over property applicability and likelihood. Thus, the property model is comparable to other mixture models implementing a \textit{prototype theory}, and it is why we call the generalized roles semantic \textit{proto}-roles.

The third component is a representation of the relationship between (a) the semantic proto-role that an argument (type) has and the syntactic positions an instantiation of that argument occupies in a particular sentence---in our case, \textit{subject}, \textit{object}, and \textit{oblique}---as well as (b) the other syntactic positions in that sentence (cf. the global model from \S\ref{sec:categoricalfeatural}). We call these instantiations \textit{argument tokens}. Because this component determines how argument tokens are linked to syntactic positions, we refer to it as the \textit{linking model}.\footnote{Unless specifically noted, this is what we mean by \textit{linking model}. When referring to the entire model, we use \textsc{sprolim}.} 

\begin{figure}[t]
\centering
\scalebox{.88}{
%
%
%
%


\begin{tikzpicture}[x=1.25cm,y=2.75cm]

  \node[obs]                   (l)      {$l$} ; %
  \node[obs, right=of l]       (a)      {$a$} ; %

  \node[latent, below=of l]    (theta)  {$\bm{\theta}$}; %
  
  \node[latent, right=of theta]    (z)  {z}; %
  \node[obs, right=of a]    (s)  {$\mathbf{s}$}; %
  \node[latent, above=of s]    (psi)  {$\psi$}; %

  \node[latent, below=.3 of s]    (c)  {$\mathbf{c}$}; %
  \node[latent, right=of z]    (phi)  {$\bm{\phi}$}; %

  \node[latent, above=of a] (eta)  {$\eta$}; %
  
  \node[latent, above=of l] (mu)  {$\mu$}; %

  \node[latent, left=.75 of l] (kappa)  {$\bm{\kappa}$}; %
  
  \node[latent, right=.75 of psi] (delta)  {$\delta$}; %
  

  \plate [xshift=1mm, yshift=-1mm, draw opacity=0] {argtokenplate'} { %
    (l)(a) %
  } {}; %
  \plate {argtokenplate} { %
    (argtokenplate') %
    (l)(a) %
  } {$\mathcal{T}_{vj}$}; %
    \plate {syntaxplate2} { %
    (delta)
  } {$\mathcal{S}$} ; %
  \plate [draw opacity=0, yshift=1.25mm] {syntaxplate2'} { %
    (delta) %
  } {}; %
  \plate {syntaxplate} { %
    (psi) %
    (syntaxplate2) %
    (syntaxplate2') %
  } {$\mathcal{S}$}; %
  \plate [draw opacity=0] {syntaxplate'} { %
    (psi) %
  } {}; %
  \plate [draw opacity=0, xshift=0mm, yshift=-2.5mm] {roleplate'} { %
    (eta) (mu) %
    (syntaxplate') %
  } {} ; %
  \plate {roleplate} { %
    (eta) (mu) %
    (syntaxplate') %
  } {$\mathcal{R}$} ; %
  \plate [draw opacity=0, inner sep=0mm, xshift=0mm, yshift=2.5mm] {propertyplate'} { %
    (argtokenplate) %
    (l) (a) %
    (mu) (eta) %
  } {} ; %
  \plate {propertyplate} { %
    (propertyplate')
    (mu) (eta) %
  } {$\mathcal{P}$} ; %
  \plate [draw opacity=0, inner sep=4mm, xshift=2.5mm, yshift=0mm] {clauseplate'} { %
    (argtokenplate) %
    (l) (a) %
    (s) %
  } {}; %
  \plate {clauseplate} { %
    (clauseplate') %
    (l) (a) %
    (s) %
    (c) %
  } {$j\in\mathcal{C}_v$}; %
    \plate [draw opacity=0, inner sep=1.75mm] {argumentplate'} { %
    (theta) (z) %
  } {}; %
  \plate {argumentplate} { %
    (theta) (z) %
  } {$\mathcal{A}_v$}; %
  \plate {argumentnumplate} { %
    (phi) %
  } {$|\mathcal{A}_v|$}; %
  \plate {verbplate} { %
    (argumentplate) %
    (clauseplate) %
    (phi)
    (theta) (s) %
  } {$v\in\mathcal{V}$} ; %

  
  \edge {theta} {z};
  \edge {phi} {c};
  \edge {eta, z, c} {a};
  \edge {kappa, mu, z, a, c} {l};
  \edge {z, psi, delta, c} {s};

\end{tikzpicture}

}
\vspace{-1mm}
\caption{Plate diagram for \textsc{sprolim} \label{fig:sprolim}}
\vspace{-3mm}
\end{figure}

The final component is a representation of the relationship between a predicate's argument tokens (in a given sentence) and that predicate's argument types. As noted by \citet{lang_unsupervised_2010}, such a component is necessary to handle argument alternations like passivization \ref{ex:passive}, the double object alternation \ref{ex:doc}, and the causative-inchoative alternation \ref{ex:causative}. If we relied purely on, e.g., relative position to associate argument tokens with argument types, we would systematically make mistakes on such cases.

\ex. \label{ex:passive}
\a. Eight Oakland players hit homers
\b. Homers were hit by eight Oakland players.

\ex. \label{ex:doc}
\a. Some 46\% give foreign cars higher quality ratings.
\b. Some 46\% give higher quality ratings to foreign cars.

\ex. \label{ex:causative}
\a. The earthquake shattered windows at SFO's air-traffic control tower.
\b. Windows at SFO's air-traffic control tower shattered.

Following \citeauthor{lang_unsupervised_2010}, we refer to this component as the \textit{canonicalizer}, and we refer to these final two components together as the \textit{mapping model}, since they define how one maps from argument tokens to argument types (labeled with semantic proto-roles), and from semantic proto-roles to syntactic positions.

In the remainder of this section, we define the property and mapping models more formally then fit \textsc{sprolim} to our data. To guide the description, Algorithm \ref{alg:sprolim} gives \textsc{sprolim}'s generative story, and Figure \ref{fig:sprolim} shows the corresponding plate diagram. 

\begin{algorithm}[t]
\caption{Semantic Proto-Role Linking Model}\label{alg:sprolim}
{\small
\begin{algorithmic}[1]
\For{verb type $v\in\mathcal{V}$}
	\For{argument type $i\in\mathcal{A}_v$}
		\State \textbf{draw} semantic protorole $z_{vi} \sim \mathrm{Cat}(\bm{\theta}_{vi})$
	\EndFor
    \For{verb token $j\in\mathcal{C}_v$}
		\State \textbf{draw} canonicalization $k \sim \mathrm{Cat}(\bm{\phi}_{v|\mathcal{T}_{vj}|})$
        \State $\cb_{vj} \gets$ element of symmetric group $S_{|\mathcal{T}_{vj}|,k}$
        \State \textbf{let} $\rb : |\mathcal{T}_{vj}|$-length tuple
      	\For{argument token $t\in\mathcal{T}_{vj}$}
        	\State $r_t \gets$ semantic protorole $z_{vc_{vjt}}$
            \For{property $p\in\mathcal{P}$}
        		\State \textbf{draw} $a_{vjt} \sim \mathrm{Bern}(\eta_{r_{vjt}p})$
                \If{$a_{vjt} = 1$}
                \State \textbf{draw} $l_{vjt} \sim \mathrm{Cat}(\mathrm{Ord}_{\bm{\kappa}}(\mu_{r_{t}p}))$
                \EndIf
        	\EndFor
        \EndFor
        \State \textbf{let} $\bm{\rho} : |\mathcal{S}^{|\mathcal{T}_{vj}|}|$-length vector
        \For{linking $\ssb'\in\mathcal{S}^{|\mathcal{T}_{vj}|}$}
        	\State $\rho_{\ssb'} \gets \prod_t \mathrm{softmax}\left(\psi_{r_t} + \sum_{o \neq t} \delta_{s'_{t}s'_{o}}\right)$
        \EndFor
        \State \textbf{draw} linking $k \sim \mathrm{Cat}(\bm{\rho})$
        \State $\ssb_{vj} \gets \mathcal{S}^{|\mathcal{T}_{vj}|}_k$ 
	\EndFor
\EndFor
\end{algorithmic}
}
\end{algorithm}

\paragraph{Property model}

The property model relates each semantic role to (i) the likelihood that a property is applicable to an argument that has that role and, (ii) if applicable, how likely it is that the property holds of that argument.

We implement this model using a cumulative link logit hurdle model \citep[see][]{agresti_categorical_2014}.\footnote{Cumulative link logit models are standard in the analysis of ordinal judgments like those contained in SPR1.} In this model, each semantic role $r\in\mathcal{R}$ is associated with two $|\mathcal{P}|$-length real-valued vectors: a real-valued vector $\bm{\mu}_r$, which corresponds to the likelihood of each property $p\in\mathcal{P}$ when an argument has role $r$, and $\bm{\eta}_k$, which gives the probability that each property $p$ is applicable to an argument that has role $r$. We first describe the cumulative link logit part of this model, which determines the probability of each likelihood rating, and then the hurdle part, which determines the probability that a particular property is applicable to a particular argument.

In the cumulative link logit portion of the model, a categorical probability mass function with support on the property likelihood ratings $l \in \{1, \ldots, 5\}$ is determined by the latent value $\mu$ and a nondecreasing real-valued cutpoint vector $\bm{\kappa}$. 

\vspace{-3mm}

\[\mathbb{P}(l=j \pipe \mu, \bm{\kappa}) = \begin{cases} 
   1 - q_{j-1} & \text{if } j = 5\\
   q_{j} - q_{j-1} & \text{otherwise} \\
  \end{cases}\]

\noindent where $q_j \equiv \mathrm{logit}^{-1}(\kappa_{j+1} - \mu)$ and $q_0 \equiv 0$. This model is known as a cumulative link logit model, since $\mathbf{q}$ is a valid cumulative distribution function for a categorical random variable with support on $\{1, \ldots, 5\}$. In Algorithm \ref{alg:sprolim}, we denote the parameters of this distribution with $\text{Ord}_{\bm{\kappa}}(\mu)$.

In the hurdle portion of the model, a Bernoulli probability mass function for applicability $a \in \{0, 1\}$ is given by $\mathbb{P}(a \pipe \eta) = \eta^a(1-\eta)^{1-a}$. What makes this model a hurdle model is that the rating probability only kicks in if the rating crosses the applicability ``hurdle.'' The procedural way of thinking about this is that, first, a rater decides whether a property is applicable; if it is not, they stop; if it is, they generate a rating. The joint probability of $l$ and $a$ is then defined as

\vspace{-4mm}

\begin{align*}
\mathbb{P}(l, a \pipe \mu, \eta, \bm{\kappa}) &= \mathbb{P}(a \pipe \eta)\mathbb{P}(l \pipe a, \mu, \bm{\kappa})\\
&\propto \mathbb{P}(a \pipe \eta)\mathbb{P}(l \pipe \mu, \bm{\kappa})^a
\end{align*}

This has the effect that the value of $\mu$ is estimated from only $(l, a)$-pairs where $a$=1---i.e., where the property was applicable to the argument.

\paragraph{Mapping model}

\begin{figure*}[t]
\hspace{-1cm}
{\scalefont{.85} \input{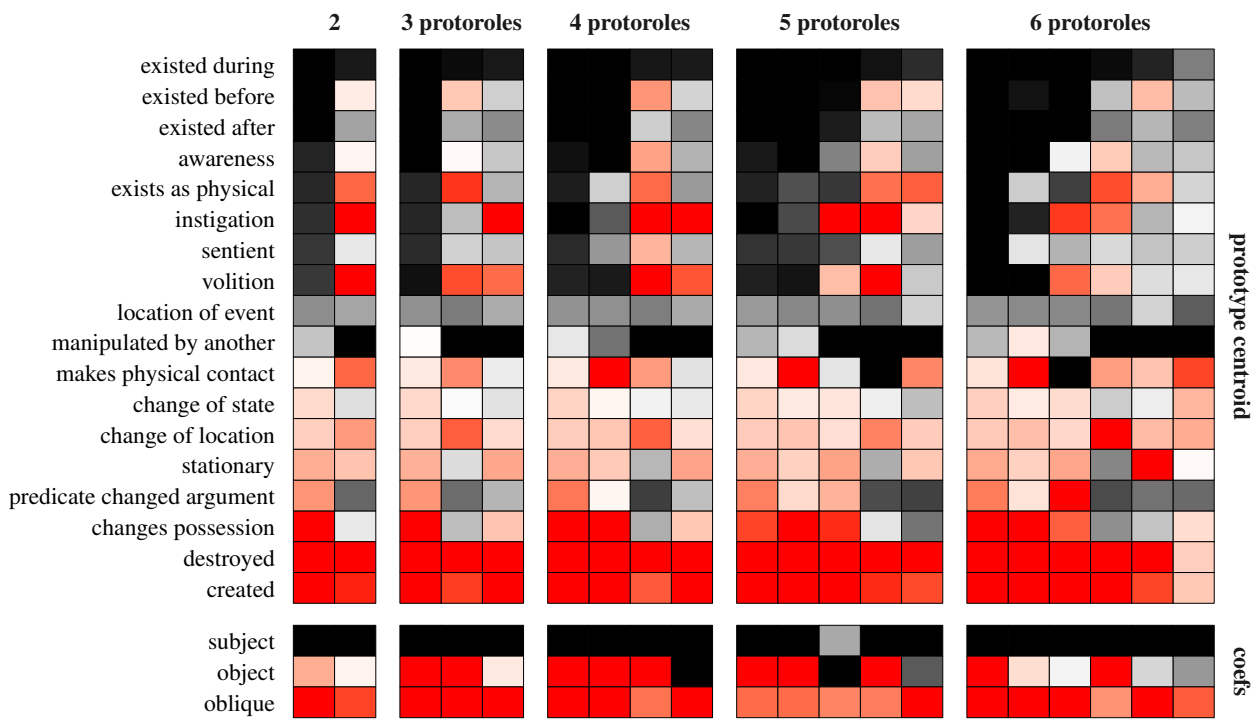}}
\vspace{-1cm}
\caption{Heatmap of prototype centroids for likelihood ratings and coefficients for each role. Black is +; red is --.\label{fig:sprolimheatmap}}
\end{figure*}

The mapping model defines how to map from argument tokens to argument types (labeled with semantic proto-roles), and from semantic proto-roles to syntactic positions. There are two components of this model: (i) the canonicalizer, which maps from argument tokens to argument types, and (ii) the linking model, which maps from argument types (labeled with semantic proto-roles) to syntactic positions.

We implement the canonicalizer by assuming that, for each predicate (verb) $v$, there is some canonical ordering of its argument types and that for each sentence (clause) $j \in \mathcal{C}_v$ that $v$ occurs in, there is some permutation of $v$'s argument tokens in that sentence that aligns them with their argument type in the canonical order. Denoting the set of argument tokens in sentence $j$ with $\mathcal{T}_{vj}$, the set of possible mappings is the symmetric group $S_{|\mathcal{T}_{vj}|}$. We place a categorical distribution with parameter $\bm{\phi}_v$ on the elements of this group.

We implement the linking model using the same CRF described in \S\ref{sec:categoricalfeatural}, but replacing the factors for each property with a factor the the arguments semantic proto-role. Thus, in Figure \ref{fig:crf}, which gives the factor graph for this CRF, the factors linking the responses $r=(a,l)$ directly to the syntactic position nodes $s$ are replaced with factors linking the semantic roles $z$ to those syntactic positions. 

\subsection{Experiment 3}
\label{sec:modelselection}

In this experiment, we fit \textsc{sprolim} to the SPR1 data and investigate the semantic protoroles it learns.

\paragraph{Model fitting} 

We use projected gradient descent with AdaGrad \citep{duchi_adaptive_2011} to find an approximation to the maximum likelihood estimates (MLE) for $\bm{\Theta}$, $\bm{\Phi}$, $\Mb$, $\Eb$, $\bm{\Psi}$, $\bm{\Delta}$, and $\bm{\kappa}$, with the variables $\Zb$ and $\Cb$ integrated out of the likelihood.

\paragraph{Determining a number of protoroles}

The one free parameter that we must set prior to fitting \textsc{sprolim} is the number of semantic protoroles $|\mathcal{R}|$.  We are interested in the model's behavior as $|\mathcal{R}|$ increases but we cannot investigate the results for all possible values here. To cut down on this set, we use a stopping criterion based on the Akaike Information Criterion (AIC). We fit \textsc{sprolim} with increasing values of $|\mathcal{R}|$, stopping when AIC is minimized. We find that the $|\mathcal{R}|$ that maximizes AIC is 6.

\paragraph{Results} 

Figure \ref{fig:sprolimheatmap} shows the estimates of the property likelihood centroids $\mathbf{L}$ (top) and the role-syntax coefficients $\bm{\Psi}$ (bottom) for each value of $|\mathcal{R}|$ fit.\footnote{The syntax-syntax coefficients show the same pattern seen among the corresponding coefficients represented in Figure \ref{fig:coefs}.} Columns give the values for a single protorole.

Perhaps the most striking aspect of this figure is that, at each value of $|\mathcal{R}|$, we see a nearly identical protorole with strong positive values on exactly \citeauthor{dowty_thematic_1991}'s \textsc{protoagent} properties and negatively (or near zero) on his \textsc{protopatient} properties. As such, we refer to this role, which is always the first column, as the \textsc{protoagent} role. 

The rest of the roles are more varied. For $|\mathcal{R}|\in\{2,3\}$, the non-\textsc{protoagent} role loads negatively (or near zero) on all \textsc{protoagent} properties, and really, all other properties besides \textsc{manipulated by another}. Comparing $|\mathcal{R}|=2$ and $|\mathcal{R}|=3$, it appears that, the non-\textsc{protoagent} role in $|\mathcal{R}|=2$ is split in two based on \textsc{instigation} and \textsc{exists as physical}, where the protorole that disprefers \textsc{instigation} is more likely to be an object.

Moving to $|\mathcal{R}|=4$, we see the addition of what looks to be a second \textsc{protoagent} role with fewer \textsc{protoagent} properties. Upon investigation of the protorole mixtures $\bm{\Theta}$ for each argument, this appears to capture cases of nonsentient or abstract---but still relatively agentive---subjects, as in \Next.

\ex. The antibody then kills the cell.

This same protorole appears in $|\mathcal{R}|=5$ with nearly all the \textsc{protoagent} properties, but dispreferring \textsc{makes physical contact}. It also appears in $|\mathcal{R}|=6$ without \textsc{sentient} or \textsc{exists as physical}, bolstering its status as capturing abstract entities---e.g., corporations.

\section{Conclusion}
\label{sec:conclusion}

In this paper, we introduced a framework for \textit{computational linking theory} (CLT) and deployed this framework for two distinct purposes: evaluation (\S\ref{sec:categoricalfeatural}) and exploration (\S\ref{sec:proto}). In \S\ref{sec:categoricalfeatural}, we evaluated four linking models based in theoretical proposals: local v. global linking models and categorical v. featural linking models. We found that global models outperform local models and categorical models outperform featural models. In \S\ref{sec:proto}, we developed the Semantic Proto-Role Linking Model in order to better understand how the property space employed in the featural models relates to the syntax. In investigating this model's behavior, we noted that it finds a protorole strikingly similar to the one proposed by \citet{dowty_thematic_1991}, but that others of \citeauthor{dowty_thematic_1991}'s protoroles fall into multiple distinct prototypes.



\bibliographystyle{plainnat} 
\bibliography{Zotero}

\end{document}